\documentclass[letterpaper, 10 pt, conference]{ieeeconf}  

\IEEEoverridecommandlockouts                              

\usepackage{amsmath} 
\usepackage{amssymb}  

\usepackage{graphicx} 
\usepackage{subcaption}
\usepackage[utf8]{inputenc}
\usepackage{pgfplots}
\pgfplotsset{compat=newest}
\usepackage{siunitx}
\usepackage[hidelinks]{hyperref}

\title{\LARGE \bf
Online Black-Box Confidence Estimation of Deep Neural Networks
}

\author{Fabian Woitschek$^{1}$ and Georg Schneider$^{1}$
\thanks{$^{1}$The authors are with ZF Friedrichshafen AG, Artificial Intelligence Lab, Saarbr\"ucken, Germany. E-Mail:
        {\tt\small firstname.surname@zf.com}}%
    \thanks{\newline \bf 978-1-6654-0540-9/22/\$31.00~\copyright2022 IEEE}
}

\begin{document}

\maketitle
\thispagestyle{empty}
\pagestyle{empty}

\begin{abstract}
Autonomous driving (AD) and advanced driver assistance systems (ADAS) increasingly utilize deep neural networks (DNNs) for improved perception or planning.
Nevertheless, DNNs are quite brittle when the data distribution during inference deviates from the data distribution during training.
This represents a challenge when deploying in partly unknown environments like in the case of ADAS.
At the same time, the standard confidence of DNNs remains high even if the classification reliability decreases.
This is problematic since following motion control algorithms consider the apparently confident prediction as reliable even though it might be considerably wrong.
To reduce this problem real-time capable confidence estimation is required that better aligns with the actual reliability of the DNN classification.
Additionally, the need exists for black-box confidence estimation to enable the homogeneous inclusion of externally developed components to an entire system.
In this work we explore this use case and introduce the neighborhood confidence (NHC) which estimates the confidence of an arbitrary DNN for classification.
The metric can be used for black-box systems since only the \mbox{top-1} class output is required and does not need access to the gradients, the training dataset or a hold-out validation dataset. 
Evaluation on different data distributions, including small in-domain distribution shifts, out-of-domain data or adversarial attacks, shows that the NHC performs better or on par with a comparable method for online white-box confidence estimation in low data regimes which is required for real-time capable AD/ADAS.
\end{abstract}

\section{INTRODUCTION} \label{sec:intro}
In recent years deep neural networks (DNNs) are increasingly used for advanced driver assistance systems (ADAS) which are deployed in public.
Further, DNNs play a key role for full autonomous driving (AD) and enable the most accurate perception, planning, etc. \cite{ad_dnn}.
At the same time, DNNs are quite brittle when the input data does not exactly match the data distribution seen during training.
Small domain shifts or the existence of out-of-domain (OOD) data lead to a significant decrease in performance.
Such shifts or small corruptions of the data occur naturally and are critical for systems deployed in partly unknown environments like ADAS.
Hence, research interest on OOD data started to increase in recent years (\cite{ood_images, ood_objects}).

The decrease in performance on OOD data is further amplified by adversarial attacks \cite{discovery}.
These attacks allow an adversary to fool any machine learning system by generating a small perturbation that is applied on the input data of the system.
Initially, such attacks applied the perturbation directly on the input image \cite{noise_physical}.
More crucial for AD/ADAS are attacks that show the possibility of performing similar attacks in the physical world.
Here, patches or markings are placed in a physical scene (\cite{patch, physical_detector, darts}) and fool the DNN even after the perturbation is captured by the camera system.
It also showed that such physical attacks are possible even if the adversary has only strict black-box access to the system \cite{physical_black_box}, meaning any deployed system could be theoretically attacked.
Hence, all described data distribution types are relevant for the safe deployment of AD/ADAS and need to be dealt with.

However, a standard DNN outputs a high confidence in the predictions on these data distributions, meaning a following planning or control algorithm treats the DNN output as a reliable value.
For a correct follow-up decision, it is required that the output confidence actually reflects the real reliability of the DNN also under data distribution shifts.
Hence, the estimated confidence should align with the true accuracy and data knowledge of the model.

This allows the following algorithms to make an informed decision and not rely on a bad prediction by the DNN.
Based on the confidence estimation the following algorithms can for example contact alternative backup systems to override the DNN prediction or engage safety features, like a speed reduction or passing the control to the (safety) driver.
Especially, the approach to use different expert models for certain situations is popular \cite{ad_fallback} and also used by current ADAS systems operating on public roads.
However, due to strict timing constraints and limited computational resources it is not possible to run the more complex expert models simultaneously.
Therefore, the general DNN must output a meaningful confidence so the controller can decide whether (and which) more complex models are used in the next images of a video stream to generate a specialized understanding of the current scene.

In reality the general DNN might be a combination of different systems from various suppliers, where each is focused on the perception of a certain task, e.g. different systems exist for driving space detection and pedestrian detection.
To still use the aforementioned concept of situational expert systems depending on the confidence it is required that the confidence estimation of each supplied system is equally meaningful.
One way to ensure this, is by using a unique and reliable confidence estimation that is independent of the individual suppliers.
This allows the final manufacturer to assess each system individually.

\begin{figure}[t]
	\centering
	\includegraphics[scale=0.23]{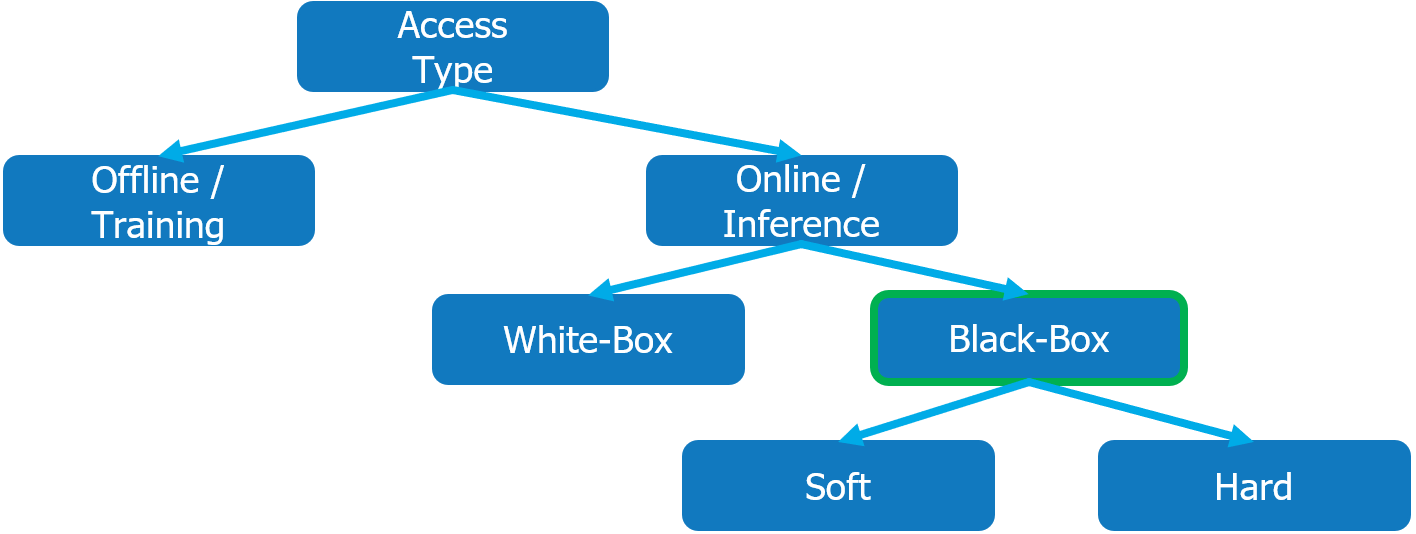}
	\caption{High-level categories to group confidence estimation methods}
	\label{fig:overview}
\end{figure}

Therefore, a model agnostic confidence estimation method is required since it is not possible to enforce a certain confidence training method for each supplier, because each has its own pipelines and architectures that cannot be easily adapted to the requirements of each customer.
Additionally, we focus on black-box confidence estimation since systems are typically shared by suppliers in a secret fashion where the customer does not have access or knowledge of the individual architecture, components or gradient flows.
Instead, only the final output of the supplied system is made available and can be used by the customer for further computations and decisions.
Hence, a model agnostic black-box confidence estimation is required to enable the safe usage of systems from different suppliers.
An overview of the different confidence estimation categories and our focus area is shown in \autoref{fig:overview}.

To explore black-box confidence estimation for ADAS we choose a traffic sign recognition (TSR) system as an exemplary ADAS, which is deployed in public by many manufacturers.
At the same time, TSR systems are most comparable to work on the confidence estimation of DNNs in the domain of image classification, where most publications are focused on.
We choose this use case since it allows to compare current advances from general image classification to our proposed black-box confidence estimation, while still being relevant for ADAS development.
This enables us to observe the difference between our black-box method and recent white-box methods.

\textbf{Our contributions:}
\begin{itemize}
	\item We motivate the need for strict black-box confidence estimation that is real-time capable for AD/ADAS
	\item The neighborhood confidence (NHC) is proposed as a method to perform black-box confidence estimation using limited additional samples
	\item A comparison with the most similar online white-box confidence method is performed for small distribution shifts, full OOD data and adversarial attacks
	\item Our findings show that we achieve an improved or similar performance in low data regimes while only using the black-box output which is required for the motivated usage in AD/ADAS
\end{itemize}

\section{RELATED WORK}
Most publications regarding the confidence estimation of DNNs use methods that change the underlying architecture or training process.
This includes the training of multiple models to use ensembles during online inference \cite{ensemble} or using different (probabilistic) layers (\cite{evidential, bayesian}) to output a more meaningful probability distribution than the standard softmax layer.
These methods are unsuited for our considered setting because they have to be applied in the training process and cannot be used to determine the reliability of a supplied system independent of the concrete supplier and system.

Alternatively, methods exist that do not require the retraining of a model and are added post hoc for inference.
Here, one method that also relies on ensembling during inference is Monte-Carlo Dropout \cite{mc_dropout} which does not require a change to the DNN architecture if dropout \cite{dropout} is already used during training.
In this approach dropout stays active during inference and allows a Bayesian approximation of the confidence.
Another method that does not require retraining is temperature scaling (\cite{calibration, calibration_bayesian}).
This improves the calibration of a trained DNN by adding a scaling coefficient to the final logit layer.
Furthermore, the combination of a DNN with the k-nearest neighbors algorithm \cite{deep_knn} is proposed to estimate the distance of a test data point to the closest train data points.
Again, the discussed methods are unsuited for the considered setting because they require an adjustment in the supplied system which a customer typically cannot perform itself.

In our considered setting model agnostic methods are needed that are only applied during inference and do not require any change in the architecture or addition of further components \cite{confidence_online}.
Here, the most similar and recently proposed method is called attribution-based confidence (ABC) \cite{attribution} which can be applied to any differentiable model and does not require any changes.
It uses the pixel-wise attribution to generate perturbed data points.
To determine the attribution gradient-based methods are exploited, meaning the ABC requires white-box access to a model.
Hence, using this method in practice requires that supplied systems are shared such that the internal computations are observable.
This contrasts with the setting we focus on, which considers secretly sharing a system and thus requires to estimate the confidence of a black-box model.

Additionally, there exists work specific to the separate data distribution types that we consider in this work.
Methods are proposed to specifically detect whether a data point is OOD \cite{ood_detection} or perturbed by an adversary \cite{adv_detection}.
However, such methods are not the focus of our work since we are interested in a single system that can provide a meaningful confidence estimation under different data distributions.
This is most useful for the scenario motivated in \autoref{sec:intro} because it allows following control algorithms to perform an appropriate action under various different conditions.
Using multiple systems to capture each data distribution type is not possible due to strict requirements on the available computational resources and timing constraints.

\section{NEIGHBORHOOD CONFIDENCE} \label{sec:nhc}
We first describe the motivation behind our proposed confidence metric.
Then, we formulate an algorithm to compute the basic version of the neighborhood confidence.
Additionally, we introduce other concepts that improve the performance of the NHC further.

\subsection{Motivation}
The basic idea behind the neighborhood confidence is that the classification reliability of a system is higher when a data point lies in the center of a decision region.
At this location the data point is classified as reliable as possible in the associated class, because a small perturbation of the data point does not change the result of the classification.
Therefore, the confidence of the system should be highest at such data points to show the highest reliability of the classification.

Following this line of thought, data points near the decision boundary are classified less reliable and should have a lower confidence.
Here, a small perturbation is sufficient to push a data point in a different decision region, which leads to a change in the classification without a meaningful change in the data itself.
Hence, the confidence of the system should be lower to show that nearly a different class is predicted and the classification is not very reliable.

\begin{figure}[t]
	\centering
	\includegraphics[scale=0.3]{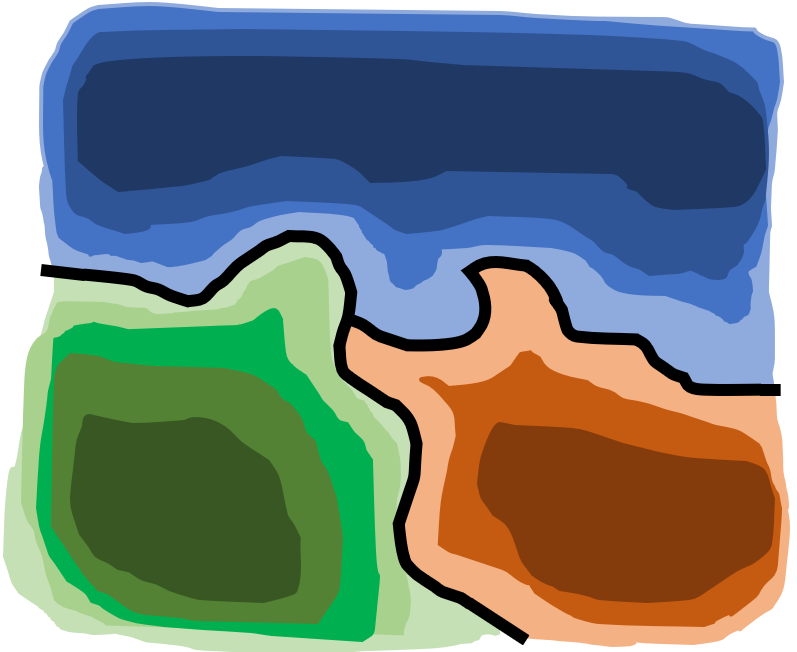}
	\caption{Simplified visualization of the classification reliability of data points in a decision region for a two dimensional example}
	\label{fig:motivation}
\end{figure}

The described basic concept behind the neighborhood confidence is visualized in \autoref{fig:motivation} for a two dimensional example with three different classes.
Darker colors show a higher value of the classification reliability and consequently an ideal confidence metric should show a similar behavior.

\subsection{Method}
Following the motivation for an ideal confidence metric, a computable black-box metric is needed that reveals how close a decision boundary is to a given data point.
Due to the high dimensionality of DNN based systems and input sensors used in reality an ideal metric can only be roughly approximated when considering the strict requirements for AD/ADAS regarding inference time and available computational resources.
To perform this approximation, we use a method that tests how robust a classification is under the influence of noise.

First, multiple noise perturbations are added on the input data and then the classification is done by the system on all data points.
If the system classifies all perturbed data points as the same class as the unperturbed data point, it shows that the input data point is not near a decision boundary.
Otherwise, the influence of the noise would have pushed some perturbed data points in a different decision region and thus a different class.
Combining the presented ideas, the final neighborhood confidence to assess the classification reliability of a black-box system is the fraction of perturbed data points that is classified as the same class as the original data point.

Hence, for a generic classification system $f(\cdot)$ the neighborhood confidence $\xi$ for a perturbation strength $\lambda$ can be calculated as:
\begin{itemize}
	\item Obtain raw data point $x \in \mathbb{R}^D$
	\item Draw $N$ noise samples $n_0, \dots , n_{N-1} \in \mathbb{R}^D$ from a random distribution
	\item Generate $N$ perturbed data points $x'_0, \dots , x'_{N-1}$ with $x'_i = x + \lambda n_i$
	\item Classify all data points $y = [f(x), f(x'_0), \dots, f(x'_{N-1})]$
	\item Calculate the NHC as $\xi = \frac{1}{N} \sum_{i=1}^{N} \{y_0 == y_i\}$
\end{itemize}

\begin{figure}[b]
	\centering
	\includegraphics[scale=0.325]{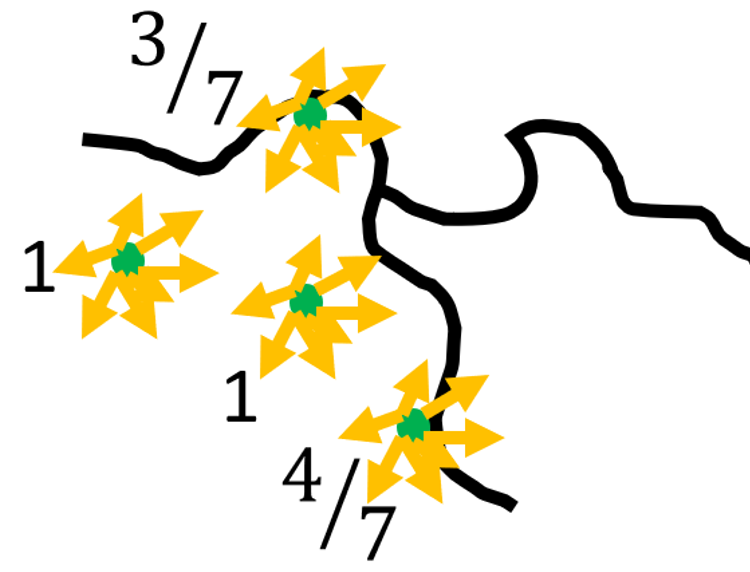}
	\caption{Simplified visualization of the resulting neighborhood confidence $\xi$ with $N=7$ for a two dimensional example}
	\label{fig:nhc}
\end{figure}

Using this method, the NHC is valued between zero and one where smaller values indicate that a decision boundary is closer.
This effect is visualized in \autoref{fig:nhc} for a simplified two dimensional example.
It can be observed that the NHC captures whether a data point is located near the boundary of a decision region or further inside that region.
The strength $\lambda$ can be used to adjust the range of the neighborhood sampling to check whether a decision boundary is nearby.
If more classes exist and the decision regions are smaller with decision boundaries closer together $\lambda$ can be decreased to still have a meaningful sampling procedure.

The described algorithm allows for an efficient calculation of the NHC, since the classification of all data points can be done in parallel.
This is beneficial for the application in environments where strict timing constraints exists, like it is the case for AD/ADAS.
If enough computational resources exist the calculation of the NHC can be done without any relevant overhead, by batching all data points $x, x'_0, \dots, x'_{N-1}$ together and calculating $y$ with a single forward pass.
To exploit this efficiency $N$ must be rather small so that the complete batch fits on the computing device and enough memory is available.
Therefore, we study the impact of the number of used noise samples in \autoref{sec:hyper} in low data regimes.

Furthermore, it is important to note that the presented NHC is model-agnostic and can be calculated without any adaption for black-box systems.
This also holds for hard black-box systems \cite{physical_black_box} where only the top-1 class is output.
No information of the classification system $f(\cdot)$ or any internal gradients is required.
This allows the application to unknown systems from external suppliers which enables the use case presented in \autoref{sec:intro}.

\subsection{Enhancements}
It is possible to enhance the presented basic version of the NHC in different ways depending on the concrete use case.
On one hand, different perturbation strengths $\lambda_1, \dots, \lambda_j$ can be used at the same time instead of only one.
This can be useful if the structure of the decision region is unknown or very uneven.
It allows to gather more insights into the structure of the surrounding decision boundaries.

Another option is to use the introduced method but specify a concrete reference class for $y_0$ instead of using the class $f(x)$ that is predicted by the system on the unperturbed data point.
This allows to estimate the distance to the decision boundary of the specified class which is useful if a potential misclassification as a certain class is more severe for some of the classes.
For instance, in the case of AD/ADAS one wants to ensure that no pedestrian detection is missed.
Hence, the pedestrian class can be chosen as reference class $y_0$, which allows to approximate the distance to this class at any time in addition to calculating the normal NHC using $f(x)$ as reference class.
If the resulting value for $\xi$ is high in the case that the pedestrian class is used as $y_0$ the decision region of the pedestrian class is close and extra care can be taken in subsequent control algorithms.

\section{EXPERIMENTS}
To evaluate the performance of the neighborhood confidence we perform qualitative experiments on different data distributions types.
The performance is compared with the most similar online white-box method ABC \cite{attribution} which mainly follows our considered setting by not requiring any adjustments or additions in the TSR system.
It is interesting to explore whether the usage of the more detailed information in the white-box method achieves a better performance than the proposed black-box method.
Since, we are interested in confidence estimation that is real-time capable we only use the gradient from a single backward pass to estimate the attribution required for the ABC, because this results in the least overhead in inference time.
Using more computational expensive methods like integrated gradients \cite{integrated_gradients} as explored in \cite{attribution} would lead to a significant delay in the time required for inference.
This goes against our goals and hence we restrict to use a single backward pass.

\subsection{Setup}
For the DNN based TSR system we choose a standard ResNet-18 architecture \cite{resnet} that is trained on the German traffic sign recognition benchmark (GTSRB) dataset \cite{gtsrb}.
Training is performed without additional augmentation and the system achieves a standard accuracy of $\approx \SI{99,3}{\percent}$ on the GTSRB final test set.

To evaluate the effect of confidence estimation under a small distribution shift in \autoref{sec:shift} we generate synthetic images of the same traffic sign classes that are used in the GTSRB dataset.
To this end, we take real images of traffic signs and apply various transformations to simulate different environmental conditions.
In this way, we generate \num{500} new samples for each of the \num{43} classes.

The performance on OOD data is evaluated in \autoref{sec:ood} by using the Chinese traffic sign recognition database (TSRD) \cite{tsrd}.
Since, we are interested in exploring the performance on full OOD data we drop all images from the TSRD that have a corresponding class in the GTSRB dataset.
Thereby, a full OOD dataset of traffic signs is generated where no overlap exists with the classes of the training dataset.

Finally, to generate adversarial attacks in \autoref{sec:adv} we use the projected gradient descent (PGD) method \cite{pgdm}.
This is a strong and standard choice to evaluate the impact of adversarial attacks and is also used by others (\cite{evidential, attribution}).

\begin{figure*}[t]
	\centering
	\begin{subfigure}[t]{0.495\linewidth}
		\centering
\begin{tikzpicture}[baseline=(current bounding box.center)]

\definecolor{color1}{rgb}{0.12156862745098,0.466666666666667,0.705882352941177}
\definecolor{color0}{rgb}{1,0.498039215686275,0.0549019607843137}
\definecolor{color2}{rgb}{0.172549019607843,0.627450980392157,0.172549019607843}
\definecolor{color3}{rgb}{0.83921568627451,0.152941176470588,0.156862745098039}

\begin{axis}[
legend cell align={left},
legend style={
  fill opacity=0.8,
  draw opacity=1,
  text opacity=1,
  at={(0.97,0.03)},
  anchor=south east,
  draw=white!80!black
},
tick align=outside,
tick pos=left,
x grid style={white!69.0196078431373!black},
xlabel={Confidence Threshold},
xmajorgrids,
xmin=-0.05, xmax=1.05,
xtick style={color=black},
y grid style={white!69.0196078431373!black},
ylabel={Accuracy / \si{\percent}},
ymajorgrids,
ymin=92.7432036217025, ymax=96.6531890605259,
ytick style={color=black},
y tick label style={xshift=.2em},
y label style={yshift=-.5em},
x tick label style={yshift=.2em},
x label style={yshift=.5em},
width=1.05\linewidth,
height=0.25\textheight,
]
\addplot [thick, color0, dashed, mark=triangle*, mark size=2, mark options={solid,rotate=180}]
table {%
0 92.9209302325581
0.05 94.7784978221608
0.1 94.7784978221608
0.15 94.7784978221608
0.2 94.7784978221608
0.25 94.7784978221608
0.3 94.7784978221608
0.35 94.7784978221608
0.4 94.7784978221608
0.45 94.7784978221608
0.5 94.7784978221608
0.55 95.6335282651072
0.6 95.6335282651072
0.65 95.6335282651072
0.7 95.6335282651072
0.75 95.6335282651072
0.8 95.6335282651072
0.85 95.6335282651072
0.9 95.6335282651072
0.95 95.6335282651072
1 95.6335282651072
};
\addlegendentry{$N = 2$}
\addplot [thick, color1, dashed, mark=triangle*, mark size=2, mark options={solid}]
table {%
0 92.9209302325581
0.05 94.5076152722721
0.1 94.5076152722721
0.15 94.5076152722721
0.2 94.5076152722721
0.25 94.8939345485556
0.3 94.8939345485556
0.35 94.8939345485556
0.4 94.8939345485556
0.45 95.2611008576702
0.5 95.2611008576702
0.55 95.2611008576702
0.6 95.2611008576702
0.65 95.6114070128717
0.7 95.6114070128717
0.75 95.6114070128717
0.8 95.6114070128717
0.85 96.1725777803243
0.9 96.1725777803243
0.95 96.1725777803243
1 96.1725777803243
};
\addlegendentry{$N = 5$}
\addplot [thick, color2, dashed, mark=triangle*, mark size=2, mark options={solid,rotate=270}]
table {%
0 92.9209302325581
0.05 94.3094065820161
0.1 94.3094065820161
0.15 94.781919111816
0.2 94.781919111816
0.25 94.781919111816
0.3 95.0842545883868
0.35 95.0842545883868
0.4 95.0842545883868
0.45 95.3072321914084
0.5 95.3072321914084
0.55 95.3072321914084
0.6 95.5561689207839
0.65 95.5561689207839
0.7 95.5561689207839
0.75 95.7905371274625
0.8 95.7905371274625
0.85 95.7905371274625
0.9 96.3376533209905
0.95 96.3376533209905
1 96.3376533209905
};
\addlegendentry{$N = 7$}
\addplot [thick, color3, dashed, mark=triangle*, mark size=2, mark options={solid,rotate=90}]
table {%
0 92.9209302325581
0.05 94.2115562879802
0.1 94.2115562879802
0.15 94.6565285789198
0.2 94.6565285789198
0.25 94.8678484237342
0.3 94.8678484237342
0.35 95.0187466523835
0.4 95.0187466523835
0.45 95.1320978983197
0.5 95.1320978983197
0.55 95.2243362590634
0.6 95.2243362590634
0.65 95.4994208813634
0.7 95.4994208813634
0.75 95.7560866651776
0.8 95.7560866651776
0.85 96.0272031737036
0.9 96.0272031737036
0.95 96.4754624496703
1 96.4754624496703
};
\addlegendentry{$N = 10$}
\end{axis}

\end{tikzpicture}
		\caption{Accuracy under consideration of confidence thresholds on the synthetic dataset}
		\label{fig:hyper_thres}
	\end{subfigure}
	\begin{subfigure}[t]{0.495\linewidth}
		\centering
\begin{tikzpicture}[baseline=(current bounding box.center)]

\definecolor{color1}{rgb}{0.12156862745098,0.466666666666667,0.705882352941177}
\definecolor{color0}{rgb}{1,0.498039215686275,0.0549019607843137}
\definecolor{color2}{rgb}{0.172549019607843,0.627450980392157,0.172549019607843}
\definecolor{color3}{rgb}{0.83921568627451,0.152941176470588,0.156862745098039}

\begin{axis}[
legend cell align={left},
legend style={
  fill opacity=0.8,
  draw opacity=1,
  text opacity=1,
  at={(0.97,0.03)},
  anchor=south east,
  draw=white!80!black
},
tick align=outside,
tick pos=left,
x grid style={white!69.0196078431373!black},
xlabel={Confidence},
xmajorgrids,
xmin=-0.0499989360570908, xmax=1.04999994933605,
xtick style={color=black},
y grid style={white!69.0196078431373!black},
ylabel={Fraction of Data},
ymajorgrids,
ymin=-0.0499802839116719, ymax=1.04958596214511,
ytick style={color=black},
ytick={1.0,0.8,0.6,0.4,0.2,0.0},
yticklabels={1.0,0.8,0.6,0.4,0.2,0.0},
y tick label style={xshift=.2em},
y label style={yshift=-.5em},
x tick label style={yshift=.2em},
x label style={yshift=.5em},
width=1.05\linewidth,
height=0.25\textheight,
]
\addplot [very thick, color0]
table {%
9.5367431640625e-07 0
9.5367431640625e-07 0.477129340171814
0.5 0.477523684501648
0.5 0.605283975601196
1 0.60567831993103
1 0.999605655670166
};
\addlegendentry{$N = 2$}
\addplot [very thick, color1, dashed]
table {%
9.5367431640625e-07 0
9.5367431640625e-07 0.403785467147827
0.200000047683716 0.404179811477661
0.200000047683716 0.479495286941528
0.399999976158142 0.479889631271362
0.399999976158142 0.541403770446777
0.600000023841858 0.541798114776611
0.600000023841858 0.600946426391602
0.799999952316284 0.601340770721436
0.799999952316284 0.668769717216492
1 0.669164061546326
1 0.999605655670166
};
\addlegendentry{$N = 5$}
\addplot [very thick, color2, dash pattern=on 1pt off 3pt on 3pt off 3pt]
table {%
9.5367431640625e-07 0
9.5367431640625e-07 0.395110368728638
0.142857074737549 0.395504713058472
0.142857074737549 0.444006323814392
0.285714268684387 0.444400668144226
0.285714268684387 0.494479537010193
0.428571462631226 0.494873762130737
0.428571462631226 0.541403770446777
0.571428537368774 0.541798114776611
0.571428537368774 0.58044171333313
0.714285731315613 0.580835938453674
0.714285731315613 0.625394344329834
0.857142925262451 0.625788688659668
0.857142925262451 0.688091516494751
1 0.688485860824585
1 0.999605655670166
};
\addlegendentry{$N = 7$}
\addplot [very thick, color3, dotted]
table {%
9.5367431640625e-07 0
9.5367431640625e-07 0.374211311340332
0.100000023841858 0.374605655670166
0.100000023841858 0.421924352645874
0.200000047683716 0.422318577766418
0.200000047683716 0.455835938453674
0.299999952316284 0.456230282783508
0.299999952316284 0.484621524810791
0.399999976158142 0.485015749931335
0.399999976158142 0.523659229278564
0.5 0.524053573608398
0.5 0.560725569725037
0.600000023841858 0.561119794845581
0.600000023841858 0.58911669254303
0.700000047683716 0.589511036872864
0.700000047683716 0.620662450790405
0.799999952316284 0.621056795120239
0.799999952316284 0.654574155807495
0.900000095367432 0.654968500137329
0.900000095367432 0.70544171333313
1 0.705835938453674
1 0.999605655670166
};
\addlegendentry{$N = 10$}
\end{axis}

\end{tikzpicture}
		\caption{Empirical CDF on the OOD dataset}
		\label{fig:hyper_ood}
	\end{subfigure}
	\caption{Comparison of the NHC for different numbers of noise samples $N$ and strength $\lambda = \num{0.2}$}
	\label{fig:hyper}
\end{figure*}
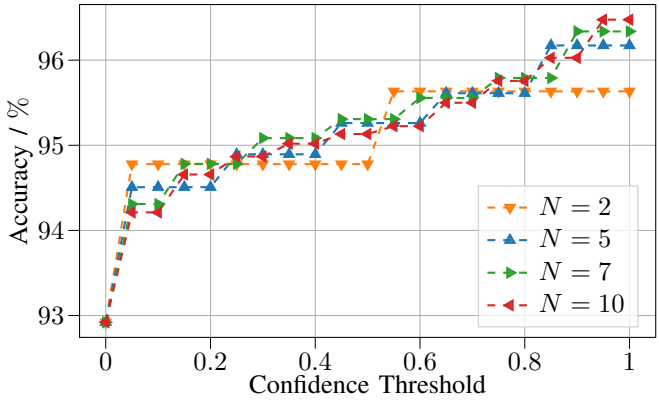
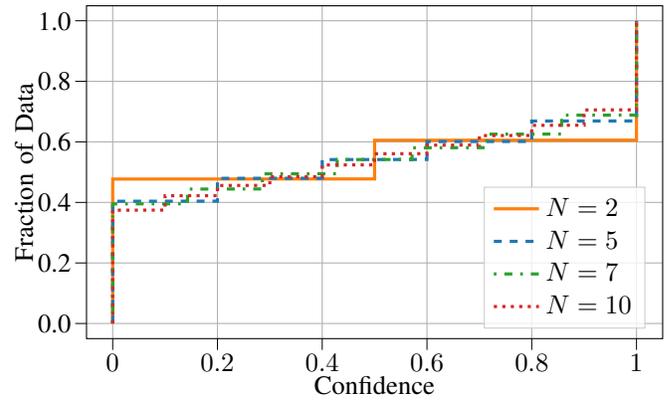

\subsection{Hyperparameter Study} \label{sec:hyper}
The limiting factor of the NHC when used for real-time capable AD/ADAS is the number of used noise samples $N$.
Hence, we visualize the impact of different choices for $N$ in \autoref{fig:hyper} for the strength $\lambda = \num{0.2}$.
However, the behavior for other strengths is very similar.
It shows, that $N$ mainly represents an option to adjust the granularity of the quantization depending on the use case.

This holds for \autoref{fig:hyper_thres} where the standard accuracy of the TSR system is shown in the case that the classification of a data point is disregarded for the calculation of the accuracy when the NHC confidence of this classification is below a threshold.
In the case that the confidence threshold equals zero the displayed value is the standard accuracy over all data points, since every classification has a confidence $\xi \geq 0$.
Thus, no classification is disregarded for the accuracy calculation.
As soon as a confidence threshold greater than zero is used the classifications of data points where the NHC is below this threshold are disregarded and the accuracy is calculated without those classifications.
For example, in \autoref{fig:hyper_thres} one can observe for $N = \num{10}$ that the accuracy is $\approx \SI{95}{\percent}$ when only classifications with a confidence $\xi \geq \num{0.4}$ are taken into account.
In this experiment one expects that the accuracy increases when the classifications with lower confidence are increasingly disregarded.

At the same time, the quantization effect also appears in \autoref{fig:hyper_ood}.
Here, cumulative distribution functions (CDFs) are used to visualize the distribution of the confidence when the classification is performed on data points of unknown classes.
For example, one can observe for $N = \num{2}$ that $\approx \SI{60}{\percent}$ of all classifications have a confidence $\xi \leq \num{0.6}$ or for $N = \num{5}$ and $N = \num{7}$ that $\approx \SI{40}{\percent}$ of all classifications have a confidence $\xi = \num{0}$.
Because we evaluate on full OOD data it is impossible that the system outputs the correct class and thus a low confidence for each classification is ideal. 

From the presented results we take that using $N = \num{7}$ achieves a good tradeoff between the quality of the confidence estimation and reduced computational requirements.
Hence, in the following we always use $N = \num{7}$ noise samples for calculating the NHC.
To have a fair comparison we also use the same number of samples for the ABC, since we are interested in a comparison under restricted computational requirements.

Additionally, we analyze the impact of the noise distribution where the samples are drawn from for the NHC.
We find that sampling from a Rademacher distribution consistently provides the best results.
A possible explanation is that this distribution allows the neighborhood sampling to search in every dimension with the maximum available strength $\lambda$ for decision boundaries.
Therefore, for the following results the noise samples are always drawn from a Rademacher distribution.

\subsection{In-Domain Distribution Shift} \label{sec:shift}
\begin{figure}[t]
	\centering
\begin{tikzpicture}[baseline=(current bounding box.center)]

\definecolor{color1}{rgb}{0.12156862745098,0.466666666666667,0.705882352941177}
\definecolor{color0}{rgb}{1,0.498039215686275,0.0549019607843137}
\definecolor{color2}{rgb}{0.172549019607843,0.627450980392157,0.172549019607843}
\definecolor{color3}{rgb}{0.83921568627451,0.152941176470588,0.156862745098039}

\begin{axis}[
legend cell align={left},
legend style={
  fill opacity=0.8,
  draw opacity=1,
  text opacity=1,
  at={(0.97,0.03)},
  anchor=south east,
  draw=white!80!black
},
tick align=outside,
tick pos=left,
x grid style={white!69.0196078431373!black},
xlabel={Confidence Threshold},
xmajorgrids,
xmin=-0.05, xmax=1.05,
xtick style={color=black},
y grid style={white!69.0196078431373!black},
ylabel={Accuracy / \si{\percent}},
ymajorgrids,
ymin=92.6975984373077, ymax=97.6108979328165,
ytick style={color=black},
y tick label style={xshift=.2em},
y label style={yshift=-.5em},
x tick label style={yshift=.2em},
x label style={yshift=.5em},
width=1.05\linewidth,
height=0.25\textheight,
ytick={93, 94, 95, 96, 97},
yticklabels={93, 94, 95, 96, 97},
]
\addplot [thick, color0, dashed, mark=triangle*, mark size=2, mark options={solid,rotate=180}]
table {%
0 92.9209302325581
0.05 95.3266402714932
0.1 95.3266402714932
0.15 95.7313013242583
0.2 95.7313013242583
0.25 95.7313013242583
0.3 96.0198624904507
0.35 96.0198624904507
0.4 96.0198624904507
0.45 96.2090887685425
0.5 96.2090887685425
0.55 96.2090887685425
0.6 96.5101214574899
0.65 96.5101214574899
0.7 96.5101214574899
0.75 96.7584406836743
0.8 96.7584406836743
0.85 96.7584406836743
0.9 97.1635001337972
0.95 97.1635001337972
1 97.1635001337972
};
\addlegendentry{$\lambda = \num{0.3}$}
\addplot [thick, color1, dashed, mark=triangle*, mark size=2, mark options={solid}]
table {%
0 92.9209302325581
0.05 95.6706092390092
0.1 95.6706092390092
0.15 96.1626644141426
0.2 96.1626644141426
0.25 96.1626644141426
0.3 96.3893850472059
0.35 96.3893850472059
0.4 96.3893850472059
0.45 96.5816530912004
0.5 96.5816530912004
0.55 96.5816530912004
0.6 96.7017543859649
0.65 96.7017543859649
0.7 96.7017543859649
0.75 97.1040453799074
0.8 97.1040453799074
0.85 97.1040453799074
0.9 97.3875661375661
0.95 97.3875661375661
1 97.3875661375661
};
\addlegendentry{$\lambda = \num{0.4}$}
\addplot [thick, color2, dashed, mark=triangle*, mark size=2, mark options={solid,rotate=270}]
table {%
0 92.9209302325581
0.05 94.8618406028774
0.1 94.8618406028774
0.15 95.2589538778665
0.2 95.2589538778665
0.25 95.2589538778665
0.3 95.4622802041974
0.35 95.4622802041974
0.4 95.4622802041974
0.45 95.8204334365325
0.5 95.8204334365325
0.55 95.8204334365325
0.6 96.1798512508452
0.65 96.1798512508452
0.7 96.1798512508452
0.75 96.4272422776331
0.8 96.4272422776331
0.85 96.4272422776331
0.9 97.1115973741794
0.95 97.1115973741794
1 97.1115973741794
};
\addlegendentry{$\lambda = \num{0.5}$}
\addplot [thick, color3, dashed, mark=triangle*, mark size=2, mark options={solid,rotate=90}]
table {%
0 92.9209302325581
0.05 94.1527446300716
0.1 94.1527446300716
0.15 94.1340184114751
0.2 94.1340184114751
0.25 94.1340184114751
0.3 94.2953776775648
0.35 94.2953776775648
0.4 94.2953776775648
0.45 94.3870361672147
0.5 94.3870361672147
0.55 94.3870361672147
0.6 94.683731233079
0.65 94.683731233079
0.7 94.683731233079
0.75 94.7722567287785
0.8 94.7722567287785
0.85 94.7722567287785
0.9 95.3213077790304
0.95 95.3213077790304
1 95.3213077790304
};
\addlegendentry{ABC}
\end{axis}

\end{tikzpicture}
	\caption{Comparison of the NHC with different strengths $\lambda$ and the ABC based on the accuracy under consideration of confidence thresholds on the synthetic dataset}
	\label{fig:shift}
\end{figure}
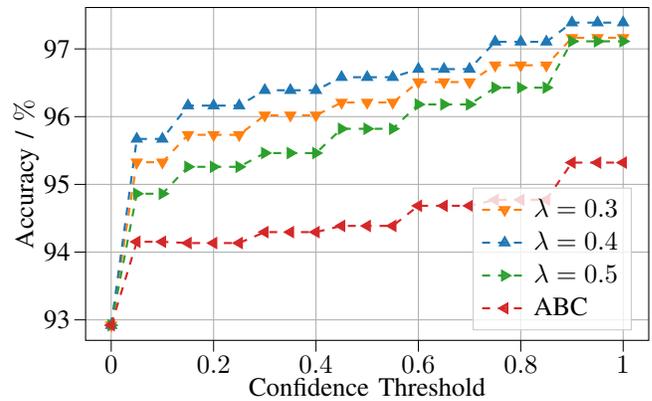

The first comparison of our proposed neighborhood confidence with the attribution-based confidence is shown in \autoref{fig:shift}.
Similar to \autoref{fig:hyper_thres}, the standard accuracy is shown when classifications of data points are disregarded for the accuracy calculation if the confidence of a classification is below a threshold.
We use our generated synthetic dataset to examine the performance under a small distribution shift where similar shifts occur naturally when deploying systems for AD/ADAS in only partly known environments.
Also, for the NHC different strengths $\lambda$ are evaluated.

All shown confidence metrics pass the basic sanity check since the accuracy increases when the confidence threshold increases.
However, all variants of the NHC reach a higher accuracy for higher confidence thresholds.
Also, they show a higher initial increase than the ABC as soon as the confidence threshold is greater than zero.
This shows that a significant fraction of the data points that are classified with $\xi = 0$ are actually misclassifications.
Once these data points are disregarded the accuracy increases notably and keeps climbing monotonously under increased confidence thresholds.

For higher confidence thresholds the NHC also achieves a higher accuracy, meaning less data points with perfect confidence are misclassified than for the ABC.
All in all, using the NHC with $\lambda = \num{0.4}$ leads to the best performance.

\subsection{Out-of-Domain Data} \label{sec:ood}
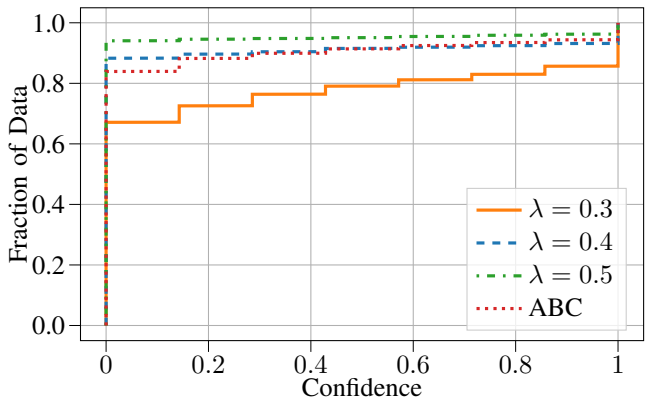
\begin{figure}[t]
	\centering
\begin{tikzpicture}[baseline=(current bounding box.center)]

\definecolor{color1}{rgb}{0.12156862745098,0.466666666666667,0.705882352941177}
\definecolor{color0}{rgb}{1,0.498039215686275,0.0549019607843137}
\definecolor{color2}{rgb}{0.172549019607843,0.627450980392157,0.172549019607843}
\definecolor{color3}{rgb}{0.83921568627451,0.152941176470588,0.156862745098039}

\begin{axis}[
legend cell align={left},
legend style={
  fill opacity=0.8,
  draw opacity=1,
  text opacity=1,
  at={(0.97,0.03)},
  anchor=south east,
  draw=white!80!black
},
tick align=outside,
tick pos=left,
x grid style={white!69.0196078431373!black},
xlabel={Confidence},
xmajorgrids,
xmin=-0.0499989360570908, xmax=1.04999994933605,
xtick style={color=black},
y grid style={white!69.0196078431373!black},
ylabel={Fraction of Data},
ymajorgrids,
ymin=-0.0499802839116719, ymax=1.04958596214511,
ytick style={color=black},
y tick label style={xshift=.2em},
y label style={yshift=-.5em},
x tick label style={yshift=.2em},
x label style={yshift=.5em},
width=1.05\linewidth,
height=0.25\textheight,
ytick={1.0,0.8,0.6,0.4,0.2,0.0},
yticklabels={1.0,0.8,0.6,0.4,0.2,0.0},
]
\addplot [very thick, color0]
table {%
9.5367431640625e-07 0
9.5367431640625e-07 0.671135663986206
0.142857074737549 0.67153000831604
0.142857074737549 0.725552082061768
0.285714268684387 0.725946426391602
0.285714268684387 0.763801336288452
0.428571462631226 0.764195561408997
0.428571462631226 0.790615081787109
0.571428537368774 0.791009426116943
0.571428537368774 0.811514139175415
0.714285731315613 0.811908483505249
0.714285731315613 0.829653024673462
0.857142925262451 0.830047369003296
0.857142925262451 0.856466889381409
1 0.856861114501953
1 0.999605655670166
};
\addlegendentry{$\lambda = \num{0.3}$}
\addplot [very thick, color1, dashed]
table {%
9.5367431640625e-07 0
9.5367431640625e-07 0.882886409759521
0.142857074737549 0.883280754089355
0.142857074737549 0.89629340171814
0.285714268684387 0.896687746047974
0.285714268684387 0.904179811477661
0.428571462631226 0.904574155807495
0.428571462631226 0.915220737457275
0.571428537368774 0.915615081787109
0.571428537368774 0.919164061546326
0.714285731315613 0.91955828666687
0.714285731315613 0.924290180206299
0.857142925262451 0.924684524536133
0.857142925262451 0.931388020515442
1 0.931782245635986
1 0.999605655670166
};
\addlegendentry{$\lambda = \num{0.4}$}
\addplot [very thick, color2, dash pattern=on 1pt off 3pt on 3pt off 3pt]
table {%
9.5367431640625e-07 0
9.5367431640625e-07 0.940457344055176
0.142857074737549 0.94085168838501
0.142857074737549 0.945583581924438
0.285714268684387 0.945977926254272
0.285714268684387 0.947949528694153
0.428571462631226 0.948343873023987
0.428571462631226 0.950709819793701
0.571428537368774 0.951104164123535
0.571428537368774 0.954653024673462
0.714285731315613 0.955047369003296
0.714285731315613 0.958596229553223
0.857142925262451 0.958990573883057
0.857142925262451 0.962145090103149
1 0.962539434432983
1 0.999605655670166
};
\addlegendentry{$\lambda = \num{0.5}$}
\addplot [very thick, color3, dotted]
table {%
9.5367431640625e-07 0
9.5367431640625e-07 0.83911669254303
0.142857074737549 0.839511036872864
0.142857074737549 0.882097721099854
0.285714268684387 0.882492065429688
0.285714268684387 0.899053573608398
0.428571462631226 0.899447917938232
0.428571462631226 0.913643598556519
0.571428537368774 0.914037942886353
0.571428537368774 0.924684524536133
0.714285731315613 0.925078868865967
0.714285731315613 0.934542655944824
0.857142925262451 0.934936881065369
0.857142925262451 0.943611979484558
1 0.944006323814392
1 0.999605655670166
};
\addlegendentry{ABC}
\end{axis}

\end{tikzpicture}
	\caption{Comparison of the NHC with different strengths $\lambda$ and the ABC based on empirical CDFs on the OOD dataset}
	\label{fig:ood}
\end{figure}

In \autoref{fig:ood} the second comparison is done by visualizing the distribution of the confidence when classifying data points of unknown classes, similar to \autoref{fig:hyper_ood}.
The OOD dataset is used which consists only of Chinese traffic signs that the TSR system trained on the GTSRB dataset cannot correctly classify.
Hence, the ideal alternative is to have a low confidence for every classification and in the best case the confidence is always zero.

A corresponding behavior can be observed for both confidence metrics.
For the NHC it shows that the strength impacts the overall confidence level.
Higher strengths lead to a decreased confidence on most data points which is intuitive.
The previously best version with $\lambda = \num{0.4}$ performs on par with the ABC and is only slightly outperformed using $\lambda = \num{0.5}$.

\subsection{Adversarial Attacks} \label{sec:adv}
\begin{figure}[b]
	\centering
\begin{tikzpicture}[baseline=(current bounding box.center)]

\definecolor{color1}{rgb}{0.12156862745098,0.466666666666667,0.705882352941177}
\definecolor{color0}{rgb}{1,0.498039215686275,0.0549019607843137}
\definecolor{color2}{rgb}{0.172549019607843,0.627450980392157,0.172549019607843}
\definecolor{color3}{rgb}{0.83921568627451,0.152941176470588,0.156862745098039}

\begin{axis}[
legend cell align={left},
legend style={fill opacity=0.8, draw opacity=1, text opacity=1, draw=white!80!black},
tick align=outside,
tick pos=left,
x grid style={white!69.0196078431373!black},
xlabel={PGD Severity $\epsilon$},
xmajorgrids,
xmin=-0.0125, xmax=0.2625,
xtick style={color=black},
y grid style={white!69.0196078431373!black},
ylabel={Confidence},
ymajorgrids,
ymin=0.0415438003211521, ymax=0.6183378632813,
ytick style={color=black},
y tick label style={xshift=.2em},
y label style={yshift=-.5em},
x tick label style={yshift=.2em},
x label style={yshift=.5em},
width=1.05\linewidth,
height=0.25\textheight,
xtick={0.0, 0.05, 0.1, 0.15, 0.2, 0.25},
xticklabels={0.0, 0.05, 0.1, 0.15, 0.2, 0.25},
ytick={0.1, 0.2, 0.3, 0.4, 0.5, 0.6},
yticklabels={0.1, 0.2, 0.3, 0.4, 0.5, 0.6},
]
\addplot [thick, color0, dashed, mark=triangle*, mark size=2, mark options={solid,rotate=180}]
table {%
0 0.592119951328566
0.025 0.283083734046581
0.05 0.170685165227846
0.075 0.178120364921038
0.1 0.203615349348201
0.125 0.229721629098404
0.15 0.258904320095861
0.175 0.285681696780892
0.2 0.313980670906777
0.225 0.342525490849517
0.25 0.371934094140696
};
\addlegendentry{$\lambda = \num{0.3}$}
\addplot [thick, color1, dashed, mark=triangle*, mark size=2, mark options={solid}]
table {%
0 0.348505580843881
0.025 0.201821356662484
0.05 0.118977593577185
0.075 0.106984246985857
0.1 0.10939620284147
0.125 0.115628752863684
0.15 0.122831400982169
0.175 0.129123749444651
0.2 0.135303143501282
0.225 0.140658619281858
0.25 0.147642001262931
};
\addlegendentry{$\lambda = \num{0.4}$}
\addplot [thick, color2, dashed, mark=triangle*, mark size=2, mark options={solid,rotate=270}]
table {%
0 0.152346324587977
0.025 0.113761655496996
0.05 0.0798746533061183
0.075 0.0705191845117613
0.1 0.0678680224307748
0.125 0.0677617122738861
0.15 0.0679942692157834
0.175 0.0679344700658044
0.2 0.068419519823651
0.225 0.0679477615356445
0.25 0.0679676941272824
};
\addlegendentry{$\lambda = \num{0.5}$}
\addplot [thick, color3, dashed, mark=triangle*, mark size=2, mark options={solid,rotate=90}]
table {%
0 0.19841938506448
0.025 0.111907844144244
0.05 0.1062932284211
0.075 0.109170297400896
0.1 0.115675273762193
0.125 0.124240048696828
0.15 0.132492528871048
0.175 0.137223423802575
0.2 0.145847998996114
0.225 0.151243340780569
0.25 0.158558948561203
};
\addlegendentry{ABC}
\end{axis}

\end{tikzpicture}
	\caption{Comparison of the NHC with different strengths $\lambda$ and the ABC under the influence of a PGD adversary}
	\label{fig:adv}
\end{figure}
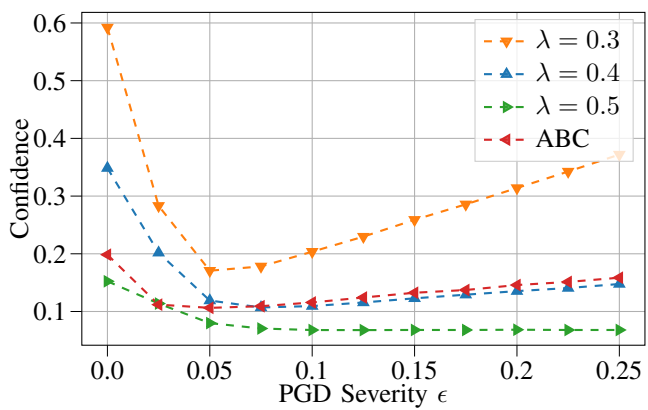

Lastly, we compare the impact of an adversary on the ABC and NHC in \autoref{fig:adv}.
The synthetic dataset is used again, and the confidence is evaluated while increasingly severe PGD attacks \cite{pgdm} are performed on the TSR system.
Here, $\epsilon$ denotes the severity of the adversary in terms of $\ell_\infty$ norm and $\epsilon = 0$ means no adversary is present.
Therefore, this is equivalent to the setting of standard classification.

The first observation is that all confidence metrics successfully decrease the confidence as soon as the adversary is introduced.
However, for the lowest strength $\lambda = \num{0.3}$ the confidence begins to significantly increase again once the severity of the adversary is further increased.
In this case, a more severe adversary can reduce the impact of the NHC because the increased severity of the adversarial perturbation pushes the data points further into the decision region of the target class of the adversary.
Using higher strengths for the neighborhood sampling prevents this effect, since the check for decision boundaries is performed at a greater distance.

Another interesting point for observation is the value of the confidence for $\epsilon = 0$.
Here, no adversary exists meaning the resulting values are the mean confidence on the unperturbed synthetic dataset.
Intuitively, the mean confidence decreases when $\lambda$ is increased for the NHC.
However, the value is also rather low for the ABC.
This means that some variants assign a low confidence to most of the classifications.
The general confidence level is sometimes low which harms the ability to correctly distinguish between benign and harmful data points when the difference to the confidence level under influence of an adversary is too low. 
In \autoref{sec:dis} we further elaborate on this behavior and the origin.

Similar to \autoref{fig:shift} the NHC with $\lambda = 0.4$ achieves the best results because the tradeoff is optimized when considering the standard mean confidence and the meaningful confidence decrease under the influence of the adversary.
This version can detect if a significant change in the distribution of the data points exists and reflects this change in the confidence.
It performs best (or close to best) on all experiments showing that a single optimal strength value can be selected which allows the efficient usage of the NHC in real applications.

\section{DISCUSSION} \label{sec:dis}
Our experiments show that for the considered low data regimes the additionally available information used in a strong gradient-based white-box method cannot be exploited and provide no benefit over the neighborhood confidence.
Instead, drawing from a Rademacher distribution provides better confidence estimates in most considered cases.
This is promising for the application in AD/ADAS since less complex methods are needed to comply with the strict timing requirements.

Our results in \autoref{sec:adv} show that the general confidence level is rather low for some evaluated variants also on unperturbed data.
The use of the synthetic dataset represents a small in-domain distribution shift that causes the data points to spread more over a decision region and lie closer to a decision boundary.
In some cases the data points also lie in a different decision region since the standard accuracy drops from $\approx \SI{99.3}{\percent}$ on the original GTSRB test dataset to $\approx \SI{92.9}{\percent}$ on our synthetic dataset (see \autoref{fig:hyper_thres} or \autoref{fig:shift}).
The reliability of the classification is reduced which is reflected in all evaluated confidence metrics.
However, for some variants the confidence level on benign data points under this distribution shift is rather low and one might want to increase the confidence gap to actually perturbed and harmful data points that are important to distinguish.
To accomplish this the integration of concepts for calibration \cite{calibration} in online confidence metrics seems promising.
It is interesting to explore the calibration of online metrics for confidence estimation depending on the current data distribution observed in past data points during inference.

\addtolength{\textheight}{-2cm}

Finally, we like to point out that it is in principle possible to combine the NHC with training methods for an improved confidence estimation.
One could explore whether the use of augmentation methods during training, like \mbox{AugMix} \cite{augmix}, has an impact on the confidence estimation.
In our preliminary experiments strong augmentation during training led to larger and more robust decision regions.
This mainly improves the behavior of all evaluated confidence metrics on unperturbed data by increasing the average confidence, while keeping the strong performance on other distribution types.
Similarly, the interaction of the NHC with adversarial training \cite{pgdm} merits a detailed investigation because adversarial training leads to increased and homogeneous decision regions around the training data samples.

\section{CONCLUSION}
We introduce the neighborhood confidence for online black-box confidence estimation of DNNs motivated by searching the neighborhood of a data point for different decision boundaries.
No information of the DNN is required and only the top-1 class output is used, which is the minimum possible output of a DNN.
This allows to use the NHC to assess the classification reliability of externally supplied components. 
The performance of the NHC is evaluated for different data distribution types deviating from the training data distribution allowing only strictly limited additional samples for inference, as required for AD/ADAS.
In this low data regime, the NHC performs better or similar to the most comparable method from the literature, even though this attribution-based confidence requires white-box access to the DNN.




%

\bibliographystyle{IEEEtran}
\bibliography{IEEEabrv,IEEEexample}

\end{document}